\pdfoutput=1
\documentclass{article}

\usepackage[preprint]{neurips_2025}

\usepackage{float}

\usepackage{graphicx}
\usepackage[utf8]{inputenc} % allow utf-8 input
\usepackage[T1]{fontenc}    % use 8-bit T1 fonts

\usepackage{url}            % simple URL typesetting
\usepackage{booktabs}       % professional-quality tables
\usepackage{amsfonts}       % blackboard math symbols
\usepackage{nicefrac}       % compact symbols for 1/2, etc.
\usepackage{microtype}      % microtypography
\usepackage{xcolor}         % colors

\usepackage[linkbordercolor=teal,citebordercolor=teal,urlbordercolor=teal,]{hyperref}
\usepackage{multirow}

\title{ Color-Style Disentangled  Two-Stream Networks for Universal Tuning-Free Style Transfer  }
\author{Shiwen Zhang \qquad Zhuowei Chen \qquad Lang Chen \qquad Yanze Wu \\
\\
\ ByteDance \vspace{+.2em}\\
\{shiwen.zhang,chenzhuowei.ustc,chenlang.cl,wuyanze.cs\}@bytedance.com
}

\begin{document}

\maketitle

\begin{abstract}
We introduce Color Disentangled Style Transfer (CDST), a novel and efficient two-stream style transfer training paradigm which completely isolates color from style and forces the style stream to be color-blinded. With one same model, CDST unlocks universal style transfer capabilities  in a tuning-free  manner during inference. Especially, the characteristics-preserved style transfer with style and content references is solved in the tuning-free way for the first time. CDST significantly improves the style similarity  by multi-feature image embeddings compression and preserves strong editing capability via our new CDST style definition inspired by Diffusion UNet disentanglement law. By conducting thorough qualitative and  quantitative experiments and human evaluations, we demonstrate that CDST  achieves state-of-the-art results on various style transfer tasks.
\end{abstract}
\section{Introduction}

\begin{figure}[!ht]
  \centering
  %\hspace{-20mm}
   \includegraphics[width=1\linewidth]{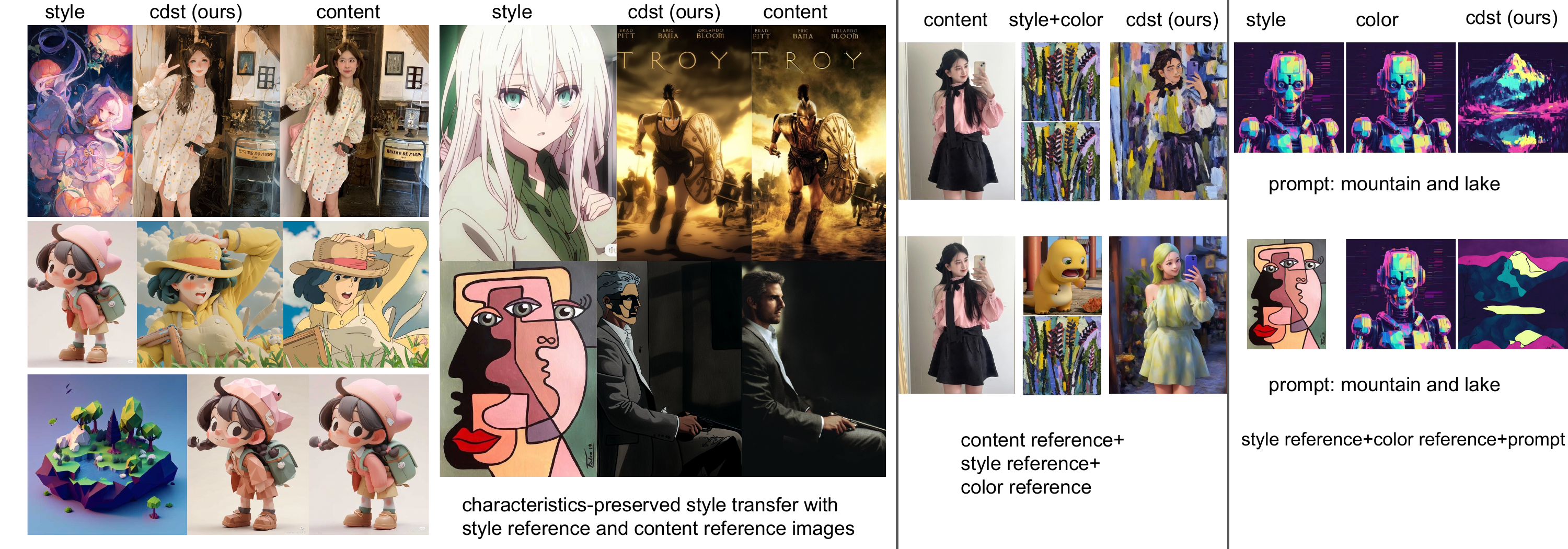}%{egfigure.eps}

   \caption{Our proposed {\bf C}olor {\bf D}isentangled {\bf S}tyle {\bf T}ransfer, CDST,    could solve all style transfer related tasks with one zero-shot style transfer model. Specifically, CDST could tackle characteristics-preserved style transfer, which transfers all style attributes except color attribute from style reference image, and preserves the color distribution of content reference image in an almost pixel-level manner so that the characteristics of content reference image could be reserved. For this task, for each example,  we put our generated results in the middle with style and content references on sides for easier visual comparison. In addition, CDST could also tackle content reference + style reference + color reference images, where style and color reference images could be identical or different. Finally, CDST solves style reference + color reference + prompt style transfer task,  where style and color reference images could be identical or different. Please zoom in for better view.  }
   \label{intro}
\end{figure}

Style transfer \citep{gatys2016image}, though a classical problem in computer vision, is actually not well-defined. It fuzzily refers to  transferring stroke, texture, material, color, light, structure etc from style reference image to content reference image or from style reference image to prompt-guided image generation. 

The cost of computation resource and time makes previous test-time tuning methods  \citep{DBLP:conf/iclr/HuSWALWWC22,sohn2023styledrop,DBLP:conf/cvpr/RuizLJPRA23,DBLP:conf/cvpr/KawarZLTCDMI23} impractical for style transfer with customized style references. Thus tuning-free  models  \citep{wang2023styleadapter} are developed for zero-shot style transfer, where the model is trained once with great amount of data and tuning-free during inference process. 

%Even for the recent Omini models based on  Diffusion Transformers, the training data is still synthesized by  SDXL style transfer models. 

However, current state-of-the-art tuning-free style transfer models still suffer from one or multiple of these problems:

1. inadequate style similarity, for example, the reconstruction of stroke, material, color, light, structure etc.

2. limited editing capability, where the model is good at reconstructing the style reference image yet fails to follow the prompt.

3. color invasion, since color is one of the main components of style,  for the content + style reference images style transfer scenario, the style model  over-transfers the color of the style reference to the content reference, leading to a significant characteristics change for the content image.  

In this paper, we introduce {\bf C}olor {\bf D}isentangled {\bf S}tyle {\bf T}ransfer, a.k.a. CDST, to tackle all the aforementioned problems of current diffusion-based tuning-free style transfer models.  Specifically, we introduce a novel two-stream color disentangled style transfer training paradigm by separating color attribute from other style concepts via quantized color histogram and greyscaled style reference.  Our CDST training paradigm costs 4 V100 GPUs with 32GB memory.  With multi-feature image embeddings compression and our new style definition via Diffusion UNet disentanglement law  and forgetting mechanisms \citep{zhang2023forgedit,fastimagic}, our CDST obtains a better style similarity and editing capability trade-off than previous state-of-the-art open-sourced and commercial models. Especially, by forcing the style stream unaware of color, our CDST unlocks the brand new capability of  characteristics-preserved style transfer.   Compatible with open-sourced SDXL ControlNet \citep{zhang2023adding}, our CDST could tackle various style-related tasks with one same style model, including but not limited to style + color + prompt, style + content + color, characteristics-preserved style + content, style+prompt, style+content, etc. Demonstrated by rich quantitative evaluations and user studies, our CDST achieves state-of-the-art results on  various  style transfer tasks in terms of style similarity, prompt alignment, color alignment, characteristics consistency, etc. 

Our main contributions are:

1. We introduce CDST, a two-stream style transfer training paradigm which completely isolates color from style, thus unlocks various new style transfer capabilities in a tuning-free manner. Especially, the characteristics-preserved style transfer with style and content references is solved for the first time.

2. CDST significantly improves the style similarity and editing capability by multi-feature image embeddings compression and our new style definition via Diffusion UNet disentanglement law. 

3. CDST  achieves state-of-the-art results on various style-related tasks through quantitative metrics and human evaluations.

Update: This model was trained in 2025.  Is CDST still useful in 2026, when FLUX \citep{flux2024,flux-2-2025}, Qwen-Image-Edit \citep{wu2025qwen} have becoming the main foundation models based on DiT \citep{dit}? Yes, CDST is used for training data synthesis in TeleStyle models \citep{qwenstyle,styleccl,telestyle,telestylev2}.

\section{Related Work}

 {\bf Style Transfer with Diffusion Models.} Diffusion Models  \citep{DBLP:conf/icml/Sohl-DicksteinW15, Ho2022ClassifierFreeDG,podell2023sdxl, Rombach2021HighResolutionIS} bring significant improvements in the domain of style transfer. Test-time tuning methods, like StyleDrop  \citep{sohn2023styledrop}, require fine-tuning the base model for every style reference image, thus are impractical for many real-world scenes. Recent methods  \citep{mou2024t2i,wang2023styleadapter,midjourney,adobe}, trained once with great amount of data, are capable of conducting tuning-free style transfer during inference.

 {\bf Disentanglement in Deep Neural Networks.} Various disentanglement properties of Deep Neural Networks \citep{he2016deep,dosovitskiy2020image,vaswani2017attention} have been explored in recent decades, for example, disentanglement in time  \citep{zhangv4d,zhang2022tfcnet,huang20204d}, in concepts  \citep{zhang2020knowledge}, in structure and texture  \citep{zhang2023forgedit,fastimagic,zhang2024hyper}. However, style is  an ill-defined  and fuzzy concept, including but not limited to color, shape, texture, lighting, structure, stroke, material etc.   Color, one of the most significant components of style, is disentangled from other style attributes by our CDST in this paper. Such disentanglement not only enables color control for style transfer tasks, but also unlocks new functionalities for example  characteristics-preserved style transfer. Previous work \citep{gatys2016preserving} tries to solve this task via image processing techniques, which is very unstable and leads to color distortion. Instruct-Imagen \citep{hu2024instruct} introduces style+subject style transfer, yet could not control whether the color of the generated subject is consistent with style reference or subject reference.
\section{CDST}
%Style has always been a complex and general concept in computer vision,  which cannot be precisely defined.  In this paper, we found that the shallow and intermediate features in Diffusion UNets are general representations for transfering colors, strokes, structures, etc, which are generally considered components of style.  Color is one of the most significant attributes among the various aspects of style concept. There are many real-world scenarios which require fine-grained controling of the color atrribute in style transfer.  In this paper, we propose a novel Color Disentangled Style Transfer (CDST) framework, which isolates color attribute from other style components thus enable us to control color and other style attributes separately, unlocking brand new applications.

In the following sections, we will first introduce the overall CDST training framework, where we will demonstrate how to disentangle color from style during training and explain how we improve the style reconstruction quality.  Then we will introduce key components in the tuning-free inference process, including our new definition of style and the global color calibration technique. Finally, we design different CDST workflows with the same trained model for unlocking different  style transfer functionalities. 
\subsection{Training Paradigm}

\begin{figure*}[!ht]
  \centering
   \includegraphics[width=1\linewidth]{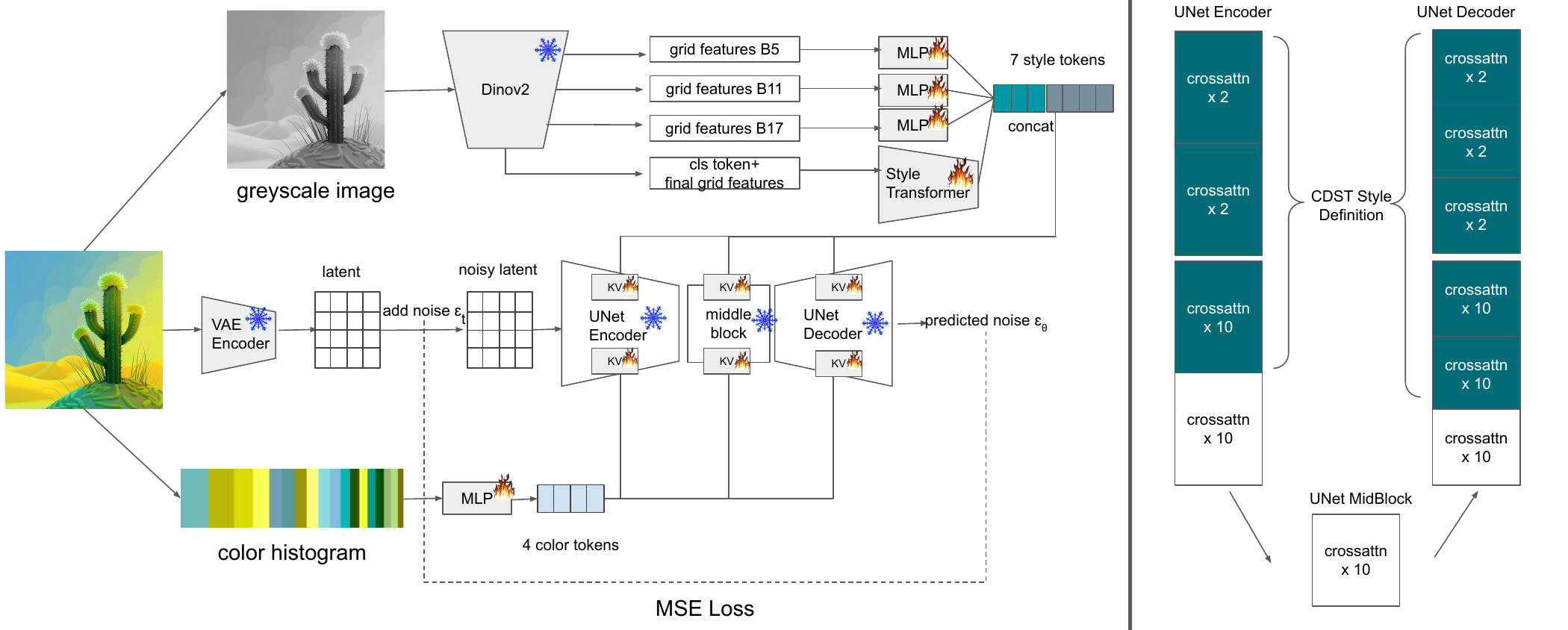}%{egfigure.eps}

   \caption{Training paradigm for CDST and our new definition of style via UNet disentanglement. We formulate the training procedure of CDST as a multi-condition image reconstruction task.   The modules with snowflake icon are frozen and the modules with fire icon are learnable.   }
   \label{training}
\end{figure*}

We formulate CDST training paradigm as a conditional reference image construction task. We propose a two-stream style transfer structure. For each training image,   we separate color attribute from other style attributes by feeding its color histogram to color stream and its greyscaled version to style stream.   We inject learnable linear matrix  $W^K_s$ and $W^V_s$ for projecting style embedding $e_s$, and $W^K_c$ and $W^V_c$ for projecting color embedding $e_c$.  We demonstrate the calculation of style stream and color stream in equation \ref{attention}, 

\begin{equation}
O=softmax(\frac{QK_t^T}{\sqrt{d}})V_t+ \lambda_s \times softmax(\frac{QK_s^T}{\sqrt{d}})V_s+\lambda_c \times softmax(\frac{QK_c^T}{\sqrt{d}})V_c
\label{attention}
\end{equation}

where  $K_s=W^K_{s}e_s$, $V_s=W^V_s e_s$, $K_c=W^K_c e_c$, $V_c=W^V_c e_c$ and $K_t,V_t$ are the projected text prompt embeddings from the original cross attention of the Denoising UNet. During training, we set $\lambda_s=1.0$ and $ \lambda_c=1.0$. The training objective is reconstructing the reference image and the training loss is the standard MSE loss. In Figure \ref{training}, we indicate the freezing parameters with snowflakes and trainable parameters with fire icon.

% The target image is disentangled into a greyscale image and color histogram and then fed to style branch and color branch separately.  Then they are fed to the greyscale style transfer branch and the  the color histogram branch.These conditions are injected into diffusion UNet via learnable cross attention blocks.

Next, we introduce how to get conditional style features $e_s$ and color features $e_c$.

\subsubsection {Style Stream with multi-feature image embeddings compression}

In order to disentangle color attribute from other style attributes, the input reference style image should not contain valid color information. Thus we feed greyscale style reference image to style stream. We first extract image features from greyscale reference image with DinoV2  \citep{oquab2023dinov2} which is composed of 50 blocks. Empirically we found that the shallow representations and the deep representations of DinoV2 play different roles for high quality image reconstruction. First, we extract deep grid features  and cls token of the last Dinov2 block. We follow Styleadapter \citep{wang2023styleadapter} which utilizes style transformer  to compress the grid embeddings into 4 tokens of 2048 dimensions.   Then,  empirically we extract shallow representations of DinoV2 from Block 5,11,17 to reconstruct  low-level cues of the reference image. The representations of each block are fed to MLPs and then average-pooled to  one token of 2048 dimensions. Thus we now have 3 tokens for shallow representations and 4 tokens for deep representations to represent greyscale style reference image.  We concat these 7 tokens to get the greyscale style embedding $e_s$, which is then mapped into Denoising UNet by new learnable cross attention blocks. We demonstrate the advantages of our style embeddings in ablation study and Figure \ref{ablation}.
\subsubsection{Color Stream with quantized color histogram}
To represent color,  we utilize the open-source package Rayleigh to extract color histogram. Rayleigh first constructs a 12x15 palette in HSV space and then the palette is  converted to LAB space. Finally, the color histogram is calculated by quantizing the pair-wise Euclidean Distance of palette LAB vector and reference image LAB vector. Eventually, the color histogram vector is a 180 dimension vector.  This vector is fed to MLP and inflated to 4 tokens of 2048 dimensions to form the color embedding $e_c$.

\subsection{Inference}

\subsubsection {New Style Definition via  UNet Disentanglement}
During training, style embeddings $e_s$ is mapped to all cross attention blocks. However, during inference, such mapping strategy will cause overfitting and content invasion, which strongly weakens the editing capablity of Diffusion UNet.  
\iffalse
Specifically, we summarize the overall structure of SDXL \\
 {[}\\
        encoder block0 2x res\\
        encoder block1 2x (res+(selfattn+crosstattn)x2)\\
        encoder block2 2x (res+(selfattn+crossattn)x10)\\
        middle block0 2xres+1x(selfattn+crossattn)x10\\
        decoder block0 3x (res+(selfattn+crossattn)x10)\\
        decoder block1 3x (res+(selfattn+crosstattn)x2)\\
        decoder block2 3x res\\   
        {]}\\ 
        \label{sdxl}
\fi
Shown in Figure \ref{training},  we denote all the cross-attention blocks with $cross\_attention\_list[0:70]$, where there are 24 cross-attention blocks in SDXL UNet Encoder, 10 cross-attention blocks in SDXL UNet Middle Blocks, and 36 cross-attention blocks in SDXL UNet Decoder, totally 70 blocks. Inspired by P+ \citep{voynov2023p+} and Forgedit  \citep{zhang2023forgedit},   we re-define the concept of style by mapping style tokens to  $cross\_attention\_list[0:14]+cross\_attention\_list[44:70]$ during inference. Please note that during training, style tokens will be mapped to all $cross\_attention\_list[0:70]$. This is because such UNet disentanglement law only appears when we project style tokens via all cross-attention blocks during training and drop deep cross-attention blocks during inference, which is called forgetting mechanism from Forgedit \citep{zhang2023forgedit}. Instead, if one only trains the CDST-style blocks, such disentanglement phenomenon will not occur, according to Forgedit \citep{zhang2023forgedit} .
   In addition, Forgedit found that in Diffusion UNet, UNet Encoder learns space and structure, UNet Decoder learns appearance and texture. Though style has no clear definition, empirically, it should be more related to appearance and texture than space and structure thus we set a very low weight $\lambda_s=0.2$  for $cross\_attention\_list[0:14]$ and  a high weight of $\lambda_s=0.9$ to $cross\_attention\_list[44:70]$. 

%we project style tokens via all cross-attention blocks during training and drop deep cross-attention blocks during inference.

%found that blocks near the middle blocks are usually unrelated to style concepts. Forgedit found that UNet encoder is related to space and structure while UNet decoder is related to appearance and texture. Thus we combine these two findings by only injecting the style embeddings to blocks in Figure \ref{} and set low style weight $\lambda_s=0.2$ in UNet encoder and high style weight $\lambda_s=0.9$ in UNet decoder. We keep a low  yet not zero value for the style weight $\lambda_s$ in UNet encoder because style is a complex concept where image structure sometimes is also a component of style. 

Our CDST style definition ensures a very strong editing capability while keeps high style similarity to the style reference image.  
\subsubsection{Color Stream}

The inference settings of color stream are very similar to its training settings. Since our CDST has already disentangled color attribute from other style attributes, we could use any color reference image  to determine the global color distribution of the generated image. The color weight is set to $\lambda_c=1.0$  for all cross-attention blocks during inference. Please note that the color stream of CDST  can only control global color distribution. The pixel-level color matching is introduced in section \ref{cpst}.

\subsubsection{Global Color Calibration}
Though the color stream controls the global color distribution well, it still has minor color approximation error due to the quantized color histogram. To remove such quantized approximation error, we further calibrate the color precision of final results by introducing Global Color Calibration during inference process.   We first convert  the generated result image and color reference image to YUV space, and then perform the following formulas, and finally convert the final result $I_{gcc}$ back to RGB space. Formally,  we denote the image generated from CDST as $I$, the color reference image as $R$, both in the YUV space,  where $H_R,W_R,H_I,W_I$ are height and width of $R$ and $I$,$\alpha$ is a hyper-parameter which controls the strength of global color calibration. We set the default value of $\alpha$ to 0.8 and further explore its effectiveness in the ablation study Figure \ref{ablation}.
\begin{equation}
\mu_R=\frac{1}{H_RW_R}\sum\limits_{h_R}^{H_R}\sum\limits_{w_R}^{W_R}R_{h_Rw_R},      \sigma_R^2=\frac{1}{H_RW_R}\sum\limits_{h_R}^{H_R}\sum\limits_{w_R}^{W_R}(R_{h_Rw_R}-\mu_R)^2
\label{eq fully connected conv}
\end{equation}

\begin{equation}
\mu_I=\frac{1}{H_IW_I}\sum\limits_{h_I}^{H_I}\sum\limits_{w_I}^{W_I}I_{h_Iw_I},      \sigma_I^2=\frac{1}{H_IW_I}\sum\limits_{h_I}^{H_I}\sum\limits_{w_I}^{W_I}(I_{h_Iw_I}-\mu_I)^2
\label{eq fully connected conv}
\end{equation}

\begin{equation}
I'=(I-\mu_I)/\sigma_I \times \sigma_R+\mu_R, I_{gcc}=\alpha I'+(1-\alpha)I
\label{eq fully connected conv}
\end{equation}

%\subsubsection{Prior Latent}
%We introduce the prior latent, which alters the inference process to an image editing task. Inspired by SDEdit, we add noise to the reference image and feed this noisy latent to Denoising UNet instead of a random Gaussian noise. This prior latent strengthen the features of reference images in different workflows which we will introduce in the next section.

\subsection{CDST Workflows}
With color completely disentangled from style reference image, we unlock several new capabilities for style transfer with one same model in tuning-free ways during inference.  It is worth noting that CDST color stream controls global color distribution and Global Color Calibration further calibrates it. The pixel-level alignment of color is introduced in the workflow of Characteristics-Preserved Content Ref + Style Ref.
\subsubsection{Characteristics-Preserved Content Ref + Style Ref}
\label{cpst}
\begin{figure*}[!ht]
  \centering
   \includegraphics[width=1\linewidth]{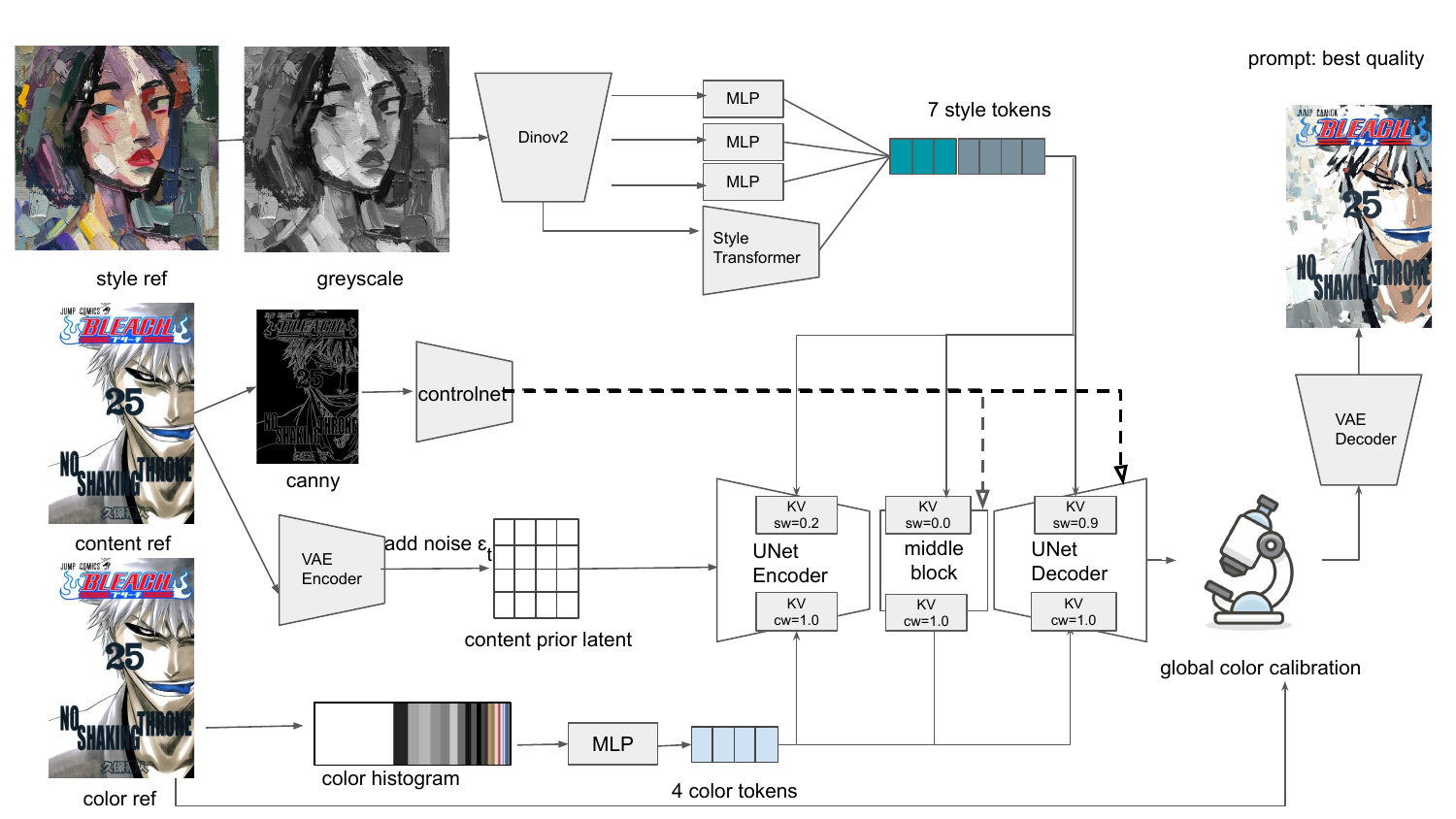}%{egfigure.eps}

   \caption{The zero-shot inference workflow of characteristics-preserved style transfer, where we transfer all style attributes except color from style reference image and keep the characteristics of the content reference image. In this workflow, content reference and color reference are the same image.}
   \label{characteristics-preserved style transfer}
\end{figure*}

%For the first time Style transfer with  content reference image and style reference image, requires the model to transfer style attributes except color from style reference image and the color of the final result should remain pixel-level aligned with content reference image to preserve the characteristics.  
For the first time, we could achieve stable and natural  characteristics-preserved style transfer given content and style reference images, which requires the model to transfer style attributes except color from style reference image and remain pixel-level aligned with content reference image to preserve the characteristics. Shown in Figure \ref{characteristics-preserved style transfer}, the pixel-level alignment is achieved via plugging in ControlNet \citep{zhang2023adding} to keep the spatial structure and Content Prior Latent to preserve the semantic pixels.  Content Prior Latent alters the inference process to an image editing task. We add noise to the content reference image and feed this noisy latent to Denoising UNet instead of a random Gaussian noise.    Given content reference image's latent $X_C$ from VAE \citep{DBLP:journals/corr/KingmaW13} and content prior strength $\lambda_P$, we  add  Gaussian Noise $\epsilon \sim \mathcal{N}(0, 1)$ to timestep $t_P=(1-\lambda_P)T$, where $T$ is the total DDIM  \citep{DBLP:conf/iclr/SongME21} sampling timesteps,
\begin{equation}
 X_{t_P}=\sqrt{\alpha _{t_P}}X_C+\sqrt{1-\alpha _{t_P}}\epsilon
\label{eq one channel normalized}
\end{equation}
with $\alpha _{t_P}$ being DDIM schedule parameter at timestep $t_P$. We call this $X_{t_P}$ Content Prior Latent, which we send to CDST via the following iterative reverse process starting from $x_t=X_{t_P}, t=t_P$, where $1<t<=t_P$, $e_t,e_s,e_c$ being text embedding, greyscale style embedding and color embedding, respectively, $I_{canny}$ being the canny image of content reference image for ControlNet :
\begin{equation}
 x_{t-1}=\frac{\sqrt{\alpha_{t-1}}}{\sqrt{\alpha_t}}(x_t-\sqrt{1-\alpha_t} \epsilon_\theta(x_t,t,e_t,e_s,e_c, I_{canny}))+\sqrt{1-\alpha_{t-1}}\epsilon_\theta(x_t,t,e_t,e_s,e_c,I_{canny})
\label{eq one channel normalized}
\end{equation}
In fact, we found that the color histogram stream does not have an significant impact in this scenario. This is because with greyscale style stream to disentangle color from style,  Content Prior Latent could  dominate the pixel-aligned color distribution. Thus we could even set color weight to 0.0 without observing obvious difference for this workflow since CDST color stream controls the global color distribution. However, forcing the style stream to be color-blinded  with greyscaled style reference is the most important  factor for preserving pixel-aligned color distribution. If the style reference is a colored image, such Content Prior Latent could not work properly, shown in ablation studies line 7 in Figure \ref{ablation}. We demonstrate the effectiveness of this workflow in Figure \ref{vis content + style}.
\subsubsection{Style Ref  + Color Ref + Content Ref}
\begin{figure*}[!ht]
  \centering
   \includegraphics[width=1\linewidth]{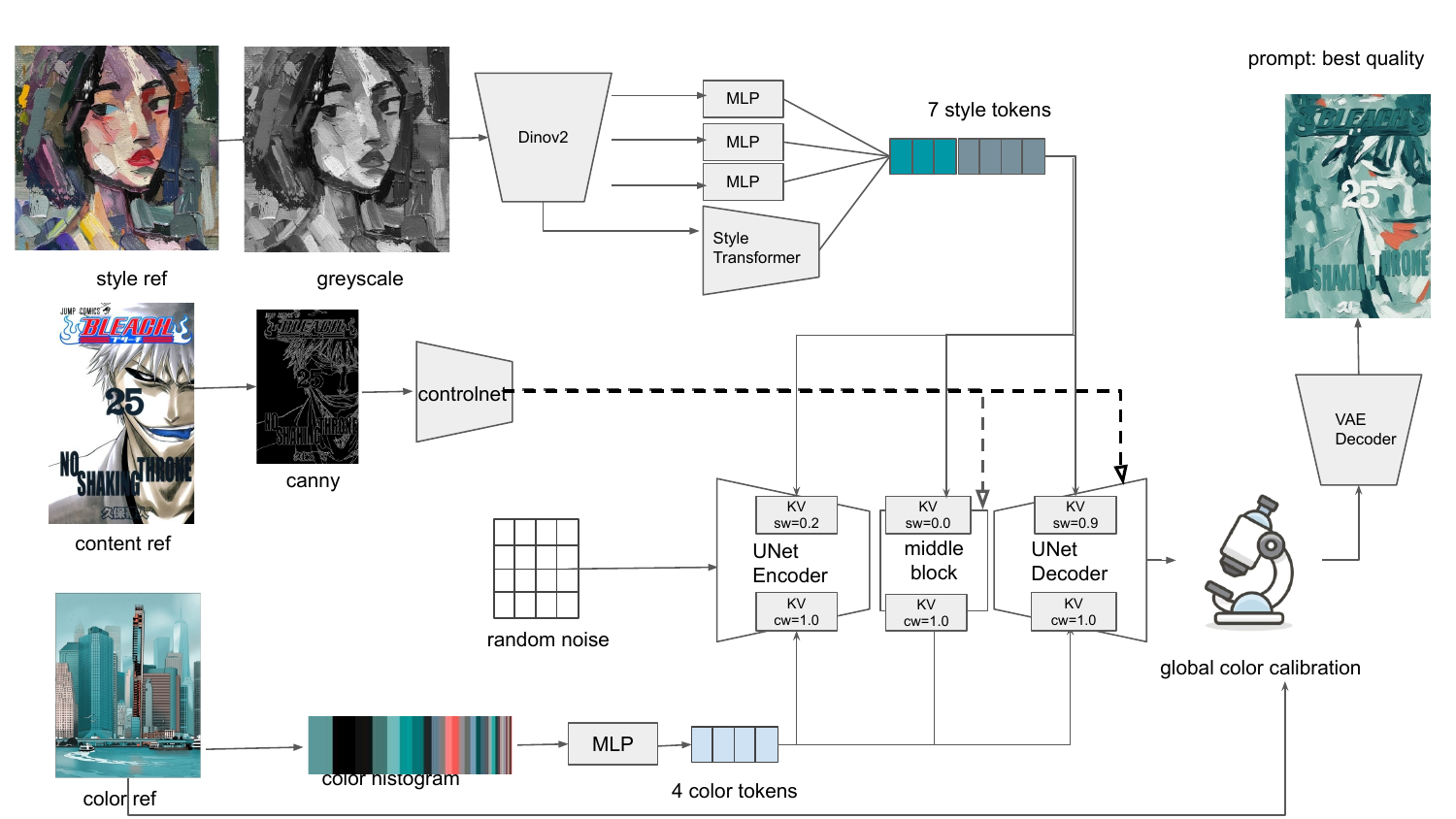}%{egfigure.eps}

   \caption{The zero-shot inference workflow of content+style+color references style transfer. The generated image has the same style of the style reference image except color, the same spatial structure of the content reference image, the same color distribution of the color reference image.  When the style and color reference images are the same, this task degrades to the traditional style reference+content reference task.  }
   \label{content+style+color}
\end{figure*}

Shown in Figure \ref{content+style+color}, we could implement tuning-free style+color+content style transfer workflow by directly incorporating open-sourced ControlNet into CDST. We demonstrate the effectiveness of this workflow in Figure \ref{vis content+style+color}.

\subsubsection{Style Ref  + Color Ref + Prompt}
\begin{figure*}[!ht]
  \centering
   \includegraphics[width=1\linewidth]{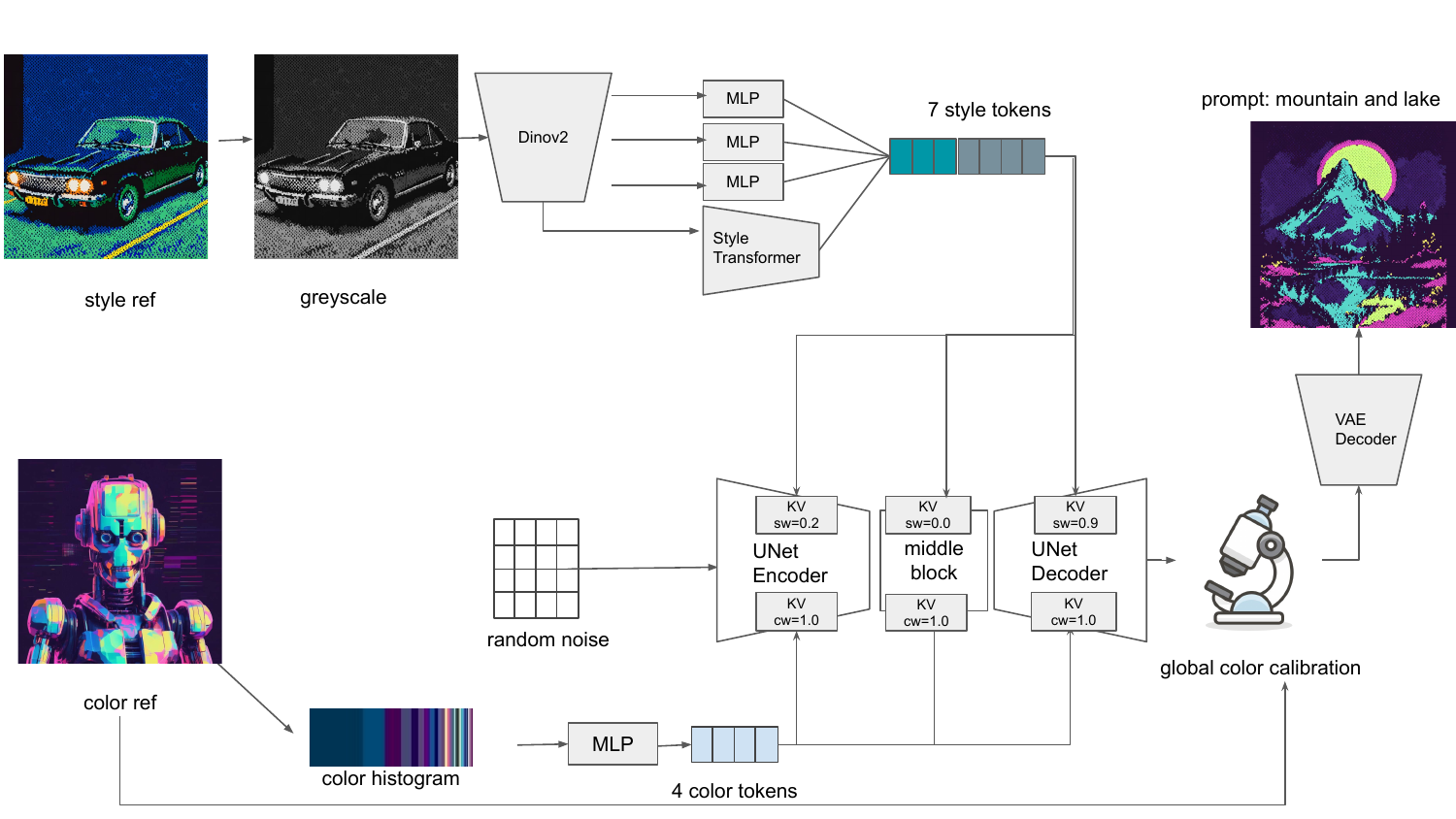}%{egfigure.eps}

   \caption{The zero-shot inference workflow of style+color+prompt, where we transfer all style attributes except color from style reference image and transfer the color distribution from color reference image. The semantics and contents of the generated result is consistent with the given prompt. When the style and color reference images are the same, this task degrades to the traditional style reference+prompt task.  }
   \label{style+prompt}
\end{figure*}

The tuning-free sampling process of style+color+prompt is shown in Figure \ref{style+prompt}. We demonstrate the effectiveness of this workflow in Figure \ref{vis style + color + prompt}.

\section{Experiments}

\subsection{Implementation Details}
{\bf Data:} CDST is trained with LAION high aesthetic data  \citep{schuhmann2022laion}, Wikiart\footnote{http://www.wikiart.org/}, and internal data.

{\bf Training:} We freeze the SDXL \citep{podell2023sdxl} base model. CDST is trained with 4 pieces of V100 of 32GB memory for 300000 iterations, with batch size of 3 on each GPU, gradient accumulation steps of 2,  learning rate of 1e-5 with adamW optimizer \citep{loshchilov2017decoupled}.

{\bf Hyper-Parameters:} We use the same hyper-parameters for each style transfer task  during inference: 

Style Ref + Color Ref+ Prompt: style weight=0.9, color weight=1.0, cfg=4.0, steps=30

Style Ref +Color Ref +Content Ref: style weight=0.9, color weight=1.0, controlnet weight=1.0, cfg=4.0, steps=30

Characteristics-Preserved Style Ref + Content Ref: style weight=0.9, color weight=1.0, controlnet weight=1.0, content prior strength=0.6, cfg=4.0, steps=30

For all these tasks, Global Color Calibration weight is set to 0.8.

\subsection{CDST Benchmark }

We collect a diverse style transfer testing benchmark, containing 25 style reference images from different genres. Since most previous solutions do not support color control, we use the same reference image for style and color to compare with previous state-of-the-art methods.  For style/color+prompt task, we randomly assign 25 prompts to 25 style references and generate 4 images with different seeds for each group. 
For characteristics-preserved style+content task, we randomly match a content reference to style reference from our collections. 

We utilize various quantitative metrics in the following experiments. For human evaluation, we invite 5 professional artists and 5 ordinary users to conduct the user study by choosing which of two given images is better in terms of a specific aspect, for example, style similarity, prompt alignment, color consistency etc.

\subsection{Ablation Studies}
\begin{figure}[H]

  \centering
   \includegraphics[width=1\linewidth]{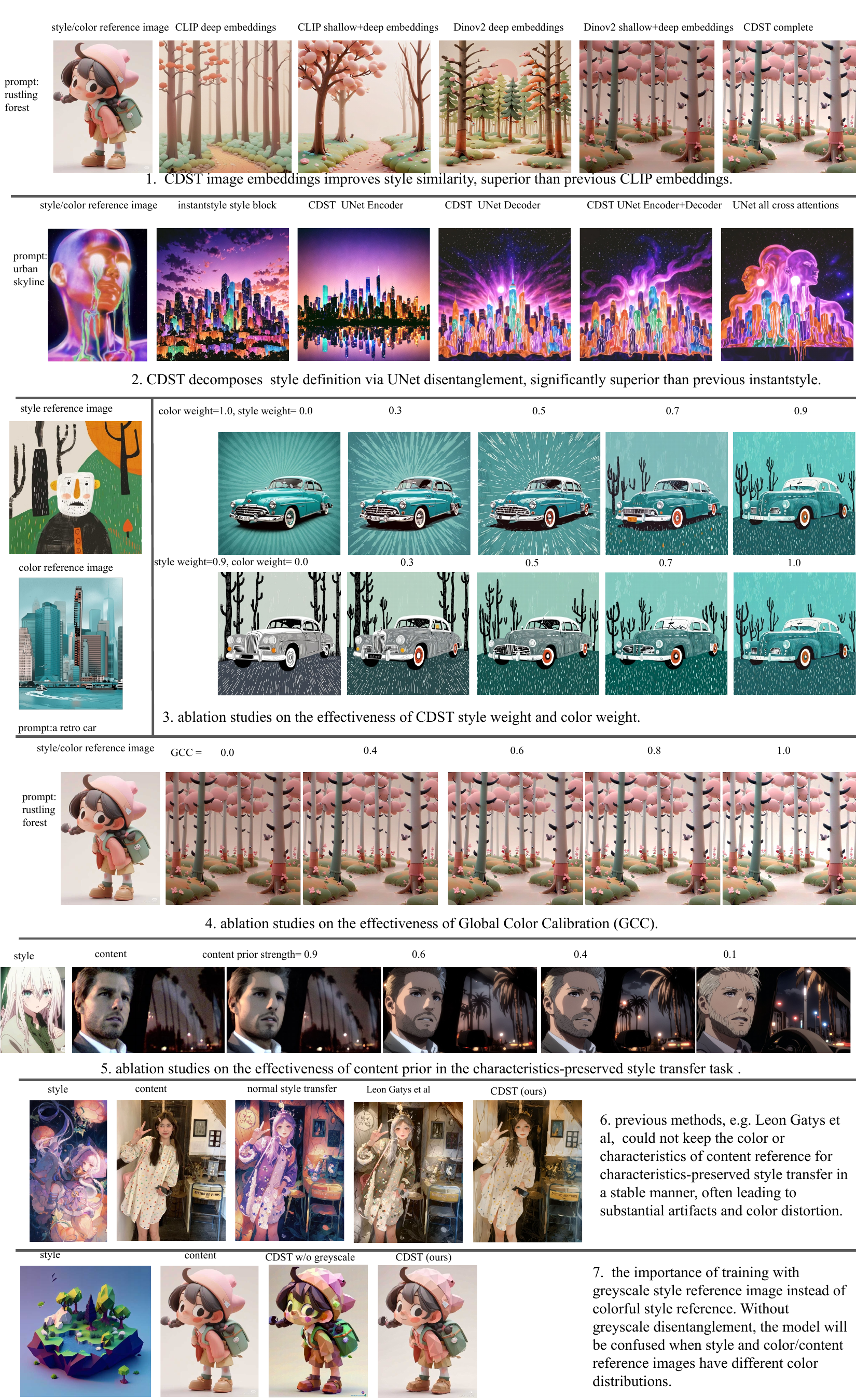}%{egfigure.eps}

   \caption{ablation studies.  }
\label{ablation}   
\end{figure}

We conduct rich ablation studies to understand and demonstrate the functionalities of CDST. We show qualitative visualizations in Figure \ref{ablation} and quantitative results in the following tables. 

{\bf Effectiveness of CDST Multi-feature Image Embedding Compression} The first line in Figure \ref{ablation} where we demonstrate that Dinov2 image embedding is better for style reconstruction than CLIP and shallow+deep image embeddings are better than deep embeddings. The  user preference is in Table \ref{ablation table}.

{\bf Superiority of CDST Style Definition. } The second line in Figure \ref{ablation} demonstrates the effectiveness of our CDST style definition. We could see that the UNet Decoder is more related to style while UNet Encoder does not contain obvious style information. However, if we combine UNet Encoder and Decoder following CDST style definition, we could observe better style transfer effects than UNet decoder only. In this example, the style of the melting effect is strengthened a little bit, which indicates that there is still spatial transformation information remaining in UNet Encoder. Instead, if we use all cross-attention blocks in UNet without forgetting mechanism, we could observe the overfitting  and content-invasion phenomenon.  The  user preference is in Table \ref{ablation table}.
 
\begin{figure}[H]
  \centering
   \includegraphics[width=1\linewidth]{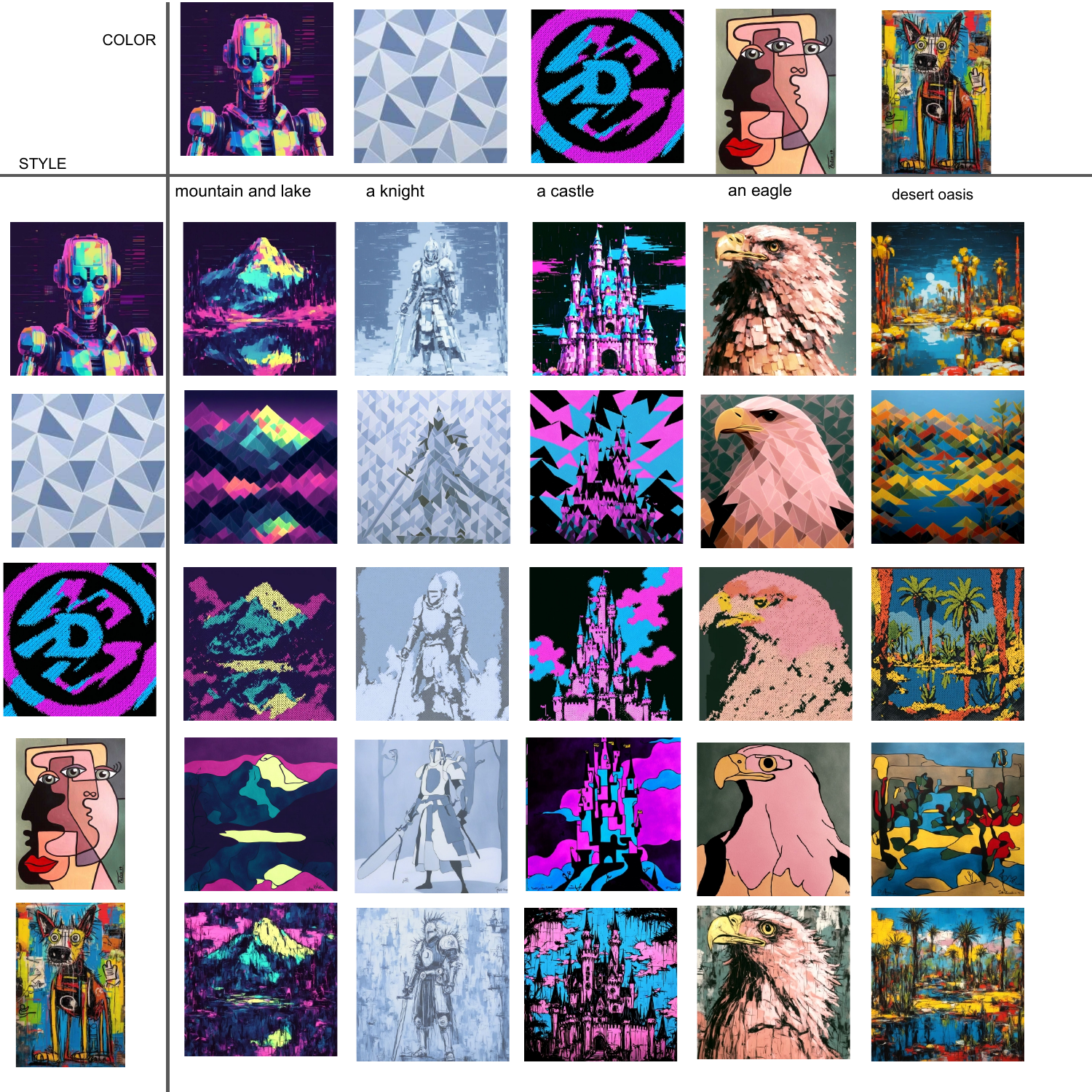}%{egfigure.eps}

   \caption{The generated results of style reference + color reference + prompt, whose inference workflow is Figure \ref{style+prompt}.    For easier visualization, we use the same prompt for each column. Thus images in each column have the same color and prompt,  and images in each row have the same style except the color attribute. Style reference and color reference are the same for  images on the  diagnose.  Please zoom in for better view. }
   \label{vis style + color + prompt}
\end{figure}

{\bf Color-Style Disentangle } Shown in Figure \ref{vis style + color + prompt}, we demonstrate that CDST completely disentangles color from style. No matter how different the color reference and style reference images are, CDST always follows the color distribution of color reference image. We further demonstrate the control by setting the color weight and style weight to different values in Figure \ref{ablation}.

\begin{figure}[t]
  \centering
   \includegraphics[width=1\linewidth]{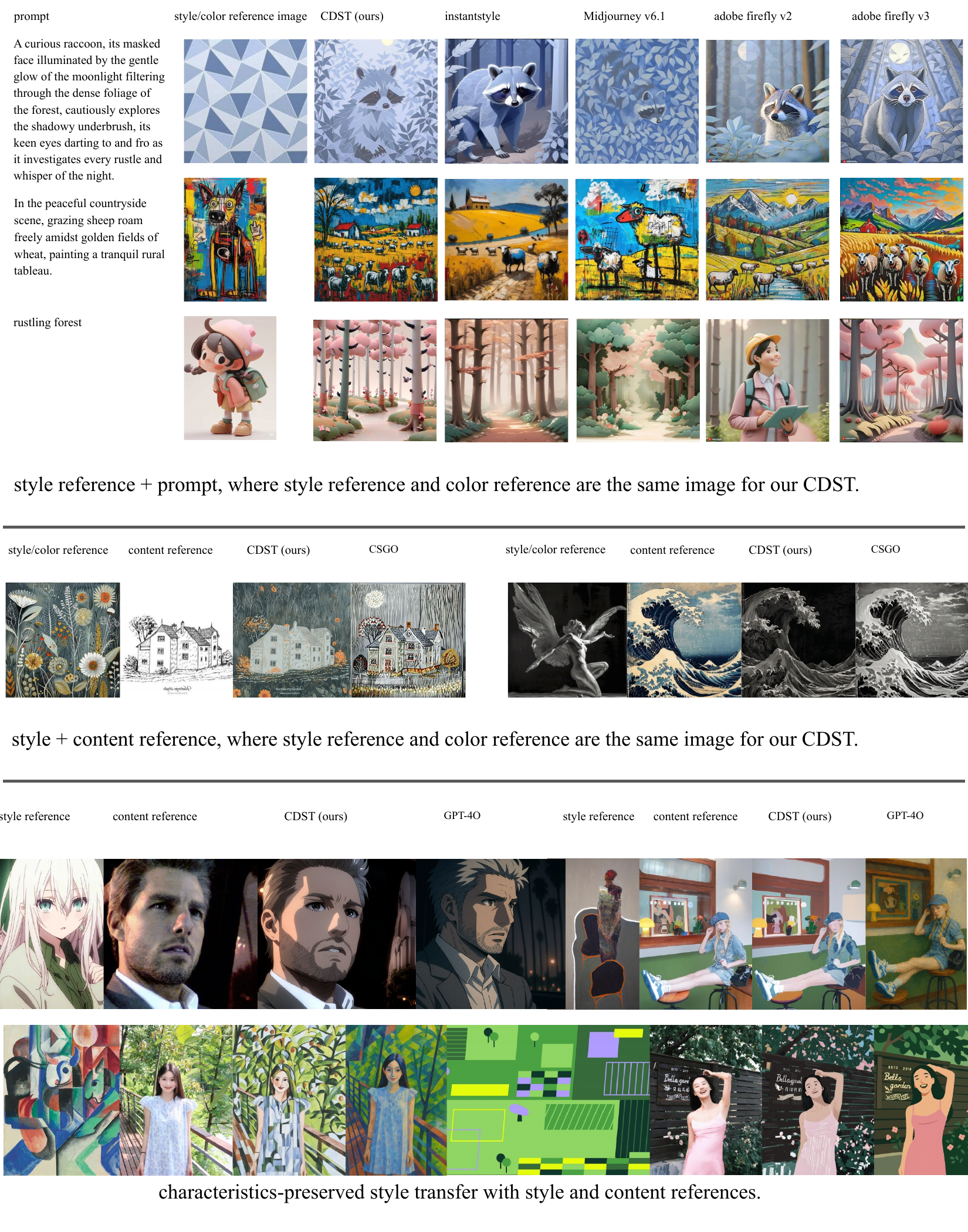}%{egfigure.eps}

   \caption{Comparison with previous state-of-the-art style transfer methods on three style tasks. Since most previous methods, either open-sourced or close-sourced, do not support color control, when we compare with such methods, we use the same reference image for style and color control.}
   \label{compare}
\end{figure}
{\bf Global Color Calibration }  To remove quantized approximation error, we demonstrate the effectiveness in Figure \ref{ablation}. When the weight of Global Color Calibration increases, we could observe the color distribution is gradually calibrated towards the reference image. 

{\bf Content Prior Strength} We study the effect of content prior strength in line 5 of Figure \ref{ablation}.

{\bf Greyscale vs Colorful Style Stream} We train a CDST model with colorful style reference for the style stream  instead of greycale image for comparison. Shown in Figure \ref{ablation}, such settings causes confusion to the network since it does not know whether it should transfer the color   from style reference or color reference. Thus  we could observe strong color distortion in the generated image.

\begin{table*}[!htb]
%\tablestyle{3pt}{1.05}
%\scriptsize
\begin{center}
\resizebox{0.97\columnwidth}{!}{
\begin{tabular}{c|c|c|c|c}
\hline
%{ } &  {CLIP deep }  & {CLIP shallow+deep} & {Dinov2 deep} & {Dinov2 shallow+deep} \\
%\hline
\multirow{2}{*}{Image Embedding}
&{CLIP deep }  & {CLIP shallow+deep} & {Dinov2 deep} & {Dinov2 shallow+deep}\\\cline{2-5}
&4.4\%& 16.0\% &16.8\%  & {\bf 62.8\%}\\

\hline
\multirow{2}{*}{Style Definition}
&{InstantStyle }  & {CDST UNet Decoder} & {CDST UNet Encoder + Decoder} & {UNet all}\\\cline{2-5}
&1.6\%& 34.8\% &{\bf 44.0\%}  & 19.6\%\\

%\hline

\hline
\end{tabular}}
\end{center}
%\vspace{-5mm}
\caption{User voting on ablation study multi-feature image embedding compression and CDST Style Definition. We invite 10 users to select the best one out of the different methods. }
\label{ablation table}

\end{table*}

\subsection{Comparison with state-of-the-art style models}
We compare our  CDST with state-of-the-art style transfer models on different style tasks. We show quantitative examples on Figure \ref{compare}. Since most previous state-of-the-art style transfer models do not support color control, we set style reference and color reference to be the same image for our CDST, for fair comparison.  For CSGO style+controlnet effects, we  compare with their reference images and results from their paper and project page.  For quantitative comparison, we employ automatic metrics and human evaluation. 

For {\bf Style/Color Ref + Prompt task}, the comparison with automatic metrics is shown in Table \ref{style+prompt compare quantitative}. We test text alignment with CLIP Score  \citep{Radford2021LearningTV}, style similarity with CSD Score  \citep{somepalli2024measuring}, aesthetic with aesthetic score  \citep{schuhmann2022improved}, color consistency with color distance. The color distance is calculated by the Euclidean Distance between color histograms of  reference image and generated image.  The user study result is shown in Table \ref{user study style+prompt}. Our CDST exhibits better performance than most previous state-of-the-art methods, except Midjourney v6.1 in terms of aesthetics.

For {\bf Style/Color Ref + Content Ref task}, we conduct  user study shown in Table \ref{user study style+content}, where our CDST wins previous state-of-the-art method CSGO in every evaluated aspects. 

For {\bf Characteristics-Preserved Style Ref + Content Ref task}, we compare with recent dominant generative model GPT-4O in Table \ref{user study characteristics-preserved style+content}, where CDST achieves better style similarity and color consistency than GPT-4O, yet performs worse than GPT-4O in terms of characteristics consistency and aesthetics. The overall user preference is 52.8\%, which indicates that our CDST and GPT-4O are almost even for this task.

\begin{table*}[!htb]
%\tablestyle{3pt}{1.05}
\scriptsize
\begin{center}
\begin{tabular}{c|c|c|c|c}
\hline
{Model} & {prompt alignment CLIP score$\uparrow$} & {style similarity CSD score$\uparrow$}  & {aesthetic score$\uparrow$} & {color distance$\downarrow$} \\
\hline
InstantStyle& \underline{0.289} & 0.390 & 6.29 & 0.176 \\
CSGO & 0.271 & 0.397 & 6.17 & 0.183\\
Midjourney v6.1 & 0.269 & 0.495 & {\bf 7.11} & 0.174\\
Adobe Firefly v2& 0.256 & 0.483 & 6.32 & 0.142\\
Adobe Firefly v3 & 0.243 & 0.441 & 5.94 & \underline{0.121}\\
StyleDrop & 0.258 & \underline{0.529} & 6.27 & 0.127\\

\hline

{\bf CDST (ours)} & {\bf 0.291} & {\bf 0.587} & \underline {6.57} & {\bf 0.116}\\

\hline
\end{tabular}
\end{center}
%\vspace{-5mm}
\caption{We compare our CDST with previous state-of-the-art style transfer methods in the style reference + prompt scenario, where color reference and style reference are the same image. We quantitatively evaluate the performance of these methods in terms of prompt alignment, style similarity, aesthetic score and color distance, where our CDST achieves the best quantitative results except aesthetic score.  The best score is stressed by bold font and the second best score is marked by underline.}
\label{style+prompt compare quantitative}
\end{table*}

\begin{table*}[!htb]
%\tablestyle{3pt}{1.05}
%\scriptsize
\begin{center}
\resizebox{0.8\columnwidth}{!}{
\begin{tabular}{c|c|c|c|c|c}
\hline
{ CDST(ours) versus } & {prompt aligment} & {style similarity }  & {aesthetic } & {color consistency} & {overall}\\
\hline
InstantStyle& 57.5\% & 90.7\% & 74.4\% & 80.4\% & 84.9\%\\
CSGO & 59.9\%  & 93.0\% & 78.3\% & 73.3\% & 84.4\% \\
Midjourney v6.1 & 69.6\% &71.1\% &{\bf 39.2}\% & 77.1\% & 57.3\% \\
Adobe Firefly v2 & 91.0\% & 87.8\% &70.5\% & 72.9\% & 83.1\%\\
Adobe Firefly v3 & 87.3\% & 89.0\%& 91.9\% &75.1\% & 88.7\%\\
StyleDrop & 67.1\% & 68.7\% & 87.9\% & 55.7\% & 78.9\%\\

\hline
\end{tabular}}
\end{center}
%\vspace{-5mm}
\caption{User Study. The number in the table represents the win rate of our CDST vs other methods. If the number is below 50\%, our CDST loses. If the number is beyond 50\%, our CDST wins. If the number is 50\%, our CDST and the compared method tie. }
\label{user study style+prompt}
\end{table*}

\begin{table*}[!htb]
%\tablestyle{3pt}{1.05}
%\scriptsize
\begin{center}
\resizebox{0.8\columnwidth}{!}{
\begin{tabular}{c|c|c|c|c|c}
\hline
{ } &  {style similarity }  & {structure consistency} & {color consistency} & {aesthetic} & {overall}\\
\hline
CDST(ours) vs CSGO & 78.0\% & 59.2\% & 81.6\% &83.2\% & 84.4\% \\

%\hline

\hline
\end{tabular}}
\end{center}
%\vspace{-5mm}
\caption{User Study on style/color + content. The number in the table represents the win rate of our CDST vs CSGO. If the number is below 50\%, our CDST loses. Otherwise, our CDST wins.  }
\label{user study style+content}
\end{table*}

\begin{table}[H]
%\tablestyle{3pt}{1.05}
%\scriptsize
\begin{center}
\resizebox{0.8\columnwidth}{!}{
\begin{tabular}{c|c|c|c|c|c}
\hline
{ } &  {style similarity }  & {characteristics consistency} & {color consistency} & {aesthetic} &{overall}\\
\hline
CDST(ours) vs GPT-4O & 63.2\% & 47.2\% & 94.8\% & 26.4\% & 52.8\% \\

%\hline

\hline
\end{tabular}}
\end{center}
%\vspace{-5mm}
\caption{User Study on characteristics-preserved style transfer. The number in the table represents the win rate of our CDST vs GPT-4O. If the number is below 50\%, our CDST loses. Otherwise, our CDST wins.  }
\label{user study characteristics-preserved style+content}
\end{table}

\subsection{Conclusion}
We propose CDST, a novel style transfer model which disentangle color from style during training.  For inference, CDST controls global color distribution via Color Stream and Global Color Calibration. The pixel-level color control is realized by Content Prior Latent and the Greyscaled Style Stream. Though CDST is trained with one style reference image, it is capable of conducting style transfer with multiple style references in a tuning-free way, either of same style or different styles. We will introduce style transfer with multiple style references in the future.

\begin{figure}[H]
  \centering
   \includegraphics[width=1\linewidth]{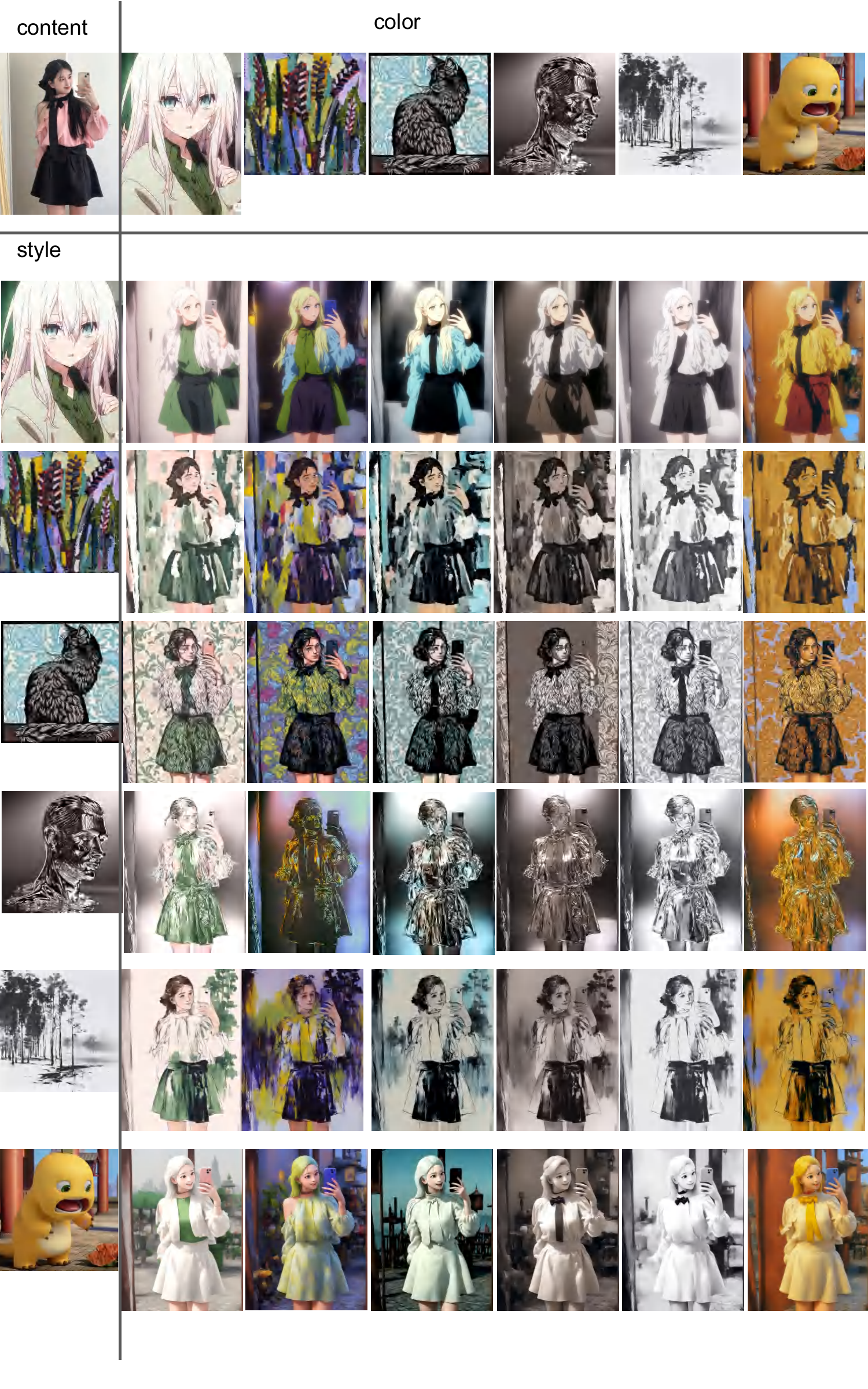}%{egfigure.eps}

   \caption{The generated results of content+style+color references, whose inference workflow is Figure \ref{content+style+color}. Style reference and color reference are the same for  images on the  diagnose. Please zoom in for better view.}
   \label{vis content+style+color}
\end{figure}
\begin{figure}[!ht]
  \centering
   \includegraphics[width=1\linewidth]{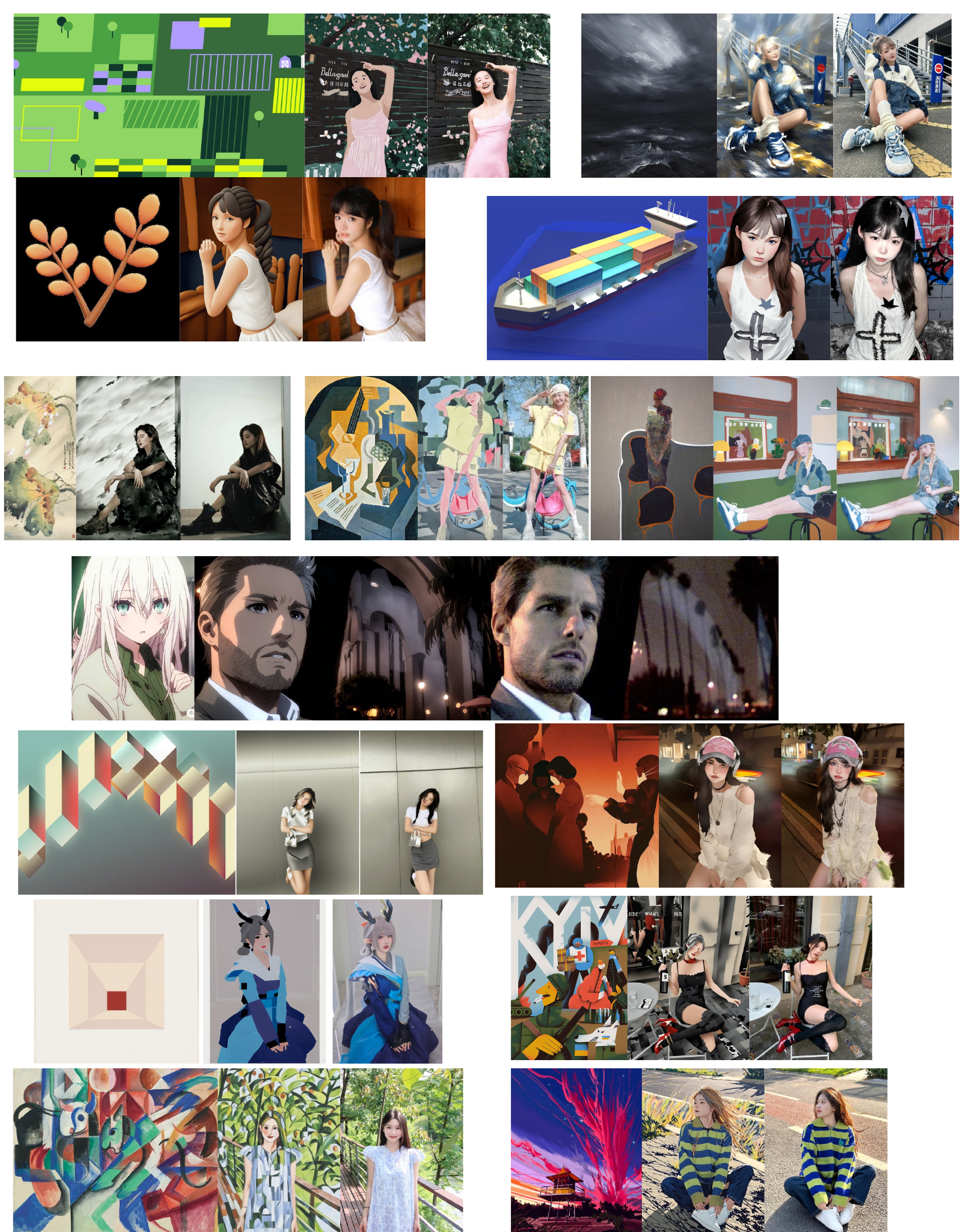}%{egfigure.eps}

   \caption{The generated results of characteristics-preserved style transfer. For easier comparison with the reference images, we put the CDST generated image in the middle for each triplet in the order of {\bf [style ref, cdst, content ref]}. Please zoom in for better view. }
   \label{vis content + style}
\end{figure}

\appendix

\section{Technical Appendices and Supplementary Material}
Due to page limit, we put generated images of CDST in the appendix.

\subsection{Style Ref + Color Ref + Prompt}
In Figure \ref{vis style + color + prompt}, we demonstrate that CDST is capable of disentangle color and style completely. Even when the colors of the style reference and color reference  are very different, the global color distribution of CDST generated results will always be consistent with color reference image. 
\subsection{Style Ref + Color Ref + Content}
In Figure \ref{vis content+style+color}, we demonstrate the color control and style transfer capbilities on the content image.

\subsection{Characteristics-Preserved Style Transfer}
In Figure \ref{vis content + style}, we demonstrate more examples of CDST on real human photos.

%%%%%%%%%%%%%%%%%%%%%%%%%%%%%%%%%%%%%%%%%%%%%%%%%%%%%%%%%%%%
\bibliography{neurips}
\bibliographystyle{plainnat}

\end{document}